\documentclass[Afour,sageh,times]{sagej}

\usepackage{moreverb,url}
\usepackage{amsmath}
\usepackage{amsfonts}
\usepackage{graphicx}
\usepackage{subfig}
\usepackage[colorlinks,bookmarksopen,bookmarksnumbered,citecolor=red,urlcolor=red]{hyperref}

\newcommand\BibTeX{{\rmfamily B\kern-.05em \textsc{i\kern-.025em b}\kern-.08em
T\kern-.1667em\lower.7ex\hbox{E}\kern-.125emX}}

\def\volumeyear{2022}

\begin{document}

%\runninghead{Bashar, Nayak and Knapman et.al.}

\title{An Informed Neural Network for Discovering Historical Documentation assisting the Repatriation of Indigenous Ancestral Human Remains}

\author{Md Abul Bashar\affilnum{1}, Richi Nayak\affilnum{1}, Gareth Knapman\affilnum{2}, Paul Turnbull\affilnum{2,3}, Cressida Fforde\affilnum{2}}

\affiliation{\affilnum{1}Queensland University of Technology, Brisbane, Australia\\
\affilnum{2}Australian National University, Canberra, Australia\\
\affilnum{3}University of Tasmania, Hobart, Australia}

%\corrauth{Richi Nayak, Queensland University of Technology,
%School of Computer Science,
%Faculty of Science,
%Brisbane, Australia.}

%\email{r.nayak@qut.edu.au}

\begin{abstract}
Among the pressing issues facing Australian and other First Nations peoples is the repatriation of the bodily remains of their ancestors, which are currently held in Western scientific institutions. The success of securing the return of these remains to their communities for reburial depends largely on locating information within scientific and other literature published between 1790-1970 documenting their theft, donation, sale, or exchange between institutions. This article reports on collaborative research by data scientists and social science researchers in the Research, Reconcile, Renew Network (RRR) to develop and apply text mining techniques to identify this vital information. We describe our work to date on developing a machine learning-based solution to automate the process of finding and semantically analyzing relevant texts. Classification models, particularly deep learning-based models, are known to have low accuracy when trained with small amounts of labelled (i.e. relevant/non-relevant) documents. To improve the accuracy of our detection model, we explore the use of an Informed Neural Network (INN) model that describes documentary content using expert-informed contextual knowledge. Only a few labelled documents are used to provide specificity to the model, using conceptually related keywords identified by RRR experts in provenance research. The results confirm the value of using an INN network model for identifying relevant documents related to the investigation of the global commercial trade in Indigenous human remains. Empirical analysis suggests that this INN model can be generalized for use by other researchers in the social sciences and humanities who want to extract relevant information from large textual corpora.
\end{abstract}


\maketitle

\section{Introduction}
\label{sec:introduction}
Text mining is a research field that has made significant progress in developing techniques for uncovering and extracting knowledge from large collections of documents. However, these techniques have mostly been applied to pre-existing textual corpora. In the humanities and social sciences, there are many pressing research questions related to past human thought and behavior for which relevant collections of historical documents do not exist in machine-readable form. Researchers must therefore use search engines to locate, gather, and analyze documentary content about past people, places, and events. This presents a challenge, as search engine results may be biased by algorithms and indexing practices, and may not accurately reflect all relevant content.

Researchers of the Research, Reconcile, Renew Network (RRR) currently face  a significant challenge as they work to assist First Nation communities in repatriating their ancestors' remains from Western scientific collections. The work involves confirming the identity, current location, and other information that is necessary for a successful reburial. To accomplish this, RRR researchers invest substantial time and intellectual energy in tracing the movement of the remains after they were stolen through as yet only partially identified networks of collectors, donors, private sellers, commercial dealers, and scientists associated with museums and other institutions with anthropological collections. Current museum catalogues provide some data for analysis, but they commonly mark the endpoints of the movement of remains through these networks, which were active from the turn of the 19th century and continue to exist today, using social media platforms, with direct messaging acting as a secure private communication channel~\citep{Huffer2019BonesPlatforms}. Moreover, relying purely on the sum of information within present-day museum catalogues runs the risk of making potentially devastating mistakes in the identification of remains~\citep{Turnbull2020Missionaries}.
 
Many relevant books and scientific journals are now available digitally (usually in PDF format). So too are ephemeral documentary sources that could be said to be the historical equivalents of modern social media, such as newspaper reportage of museum donations and exhibitions, advertisements for auctions and accounts of the sale of private anthropological collections. All of these sources have proven invaluable in reconstructing the fate of ancestral remains in historically active networks~\citep{Knapman2020ProvenanceMedia}. However, the practicality of finding and thoroughly investigating these diverse sources for potentially crucial information for repatriating communities is currently severely constrained by the fact that they are distributed across many different online locations. 

Digital library initiatives by national and major research libraries since the mid-1990s have made it possible to systematically search large historical corpora of books, journals and newspapers. However, finding relevant information about the scientific theft and uses of the ancestral remains of Australian and other First Nation peoples within these corpora is difficult.  Often these corpora comprise publications, the content of which covers a wide spectrum of topics. This is especially true of newspapers.  For RRR researchers, the usefulness of keyword-based search services provided by the creators of corpora has proven to greatly depend on how queries are formed. If the search terms used are too specific, few or no relevant documents will be returned. If terms are generic, large numbers of irrelevant documents will be returned.  Successful searching thus relies heavily on expert ‘human’ users impractically investing time and intellectual energy in searching and filtering results - when pressing needs of repatriating communities whom they are assisting need addressing in relatively short time-frames.   Assessing whether documents returned by a search engine are relevant to the needs of the user is consequently the foremost text mining problem to be solved.
%Reporting of human remains collections and the findings of the scientists who examined them are found in diverse publications including newspapers, scientific journals, naturalist societies' transactions, medical journals, and so on. 

The use of machine learning to automate the process of identifying relevant documents is the obvious solution, but the text classification task involved is challenging due to the historical nature of the documents. There is a high risk of noise and error occurring during the process of reproducing them as OCRed text due to their physical condition. Even if good digital transcripts are obtained, there is the need to account for linguistic changes, shifts in conceptual vocabularies, and the use of different medical and scientific nomenclatures, as well as the possibility of texts comprising several languages. Existing language models, such as BERT~\citep{Devlin2018BERT:Understanding} and RoBERTa~\citep{Liu2019RoBERTa:Approach}, are likely to perform poorly in recognizing the semantics and contextual relations within these documents because they are semantically and structurally different from the modern documents that are used to pre-train these models. Furthermore and most importantly, there is no labelled dataset to effectively train classification models, which is a fundamental necessity for building a text classifier.

In this paper, we describe work on designing a supervised learning model, namely an Informed Neural Network (INN) to automate the detection of relevant documents.  A set of keywords (e.g. as given in Table \ref{tab:R3Keywords}) has been provided by RRR researchers. Based on their expert knowledge they have chosen keywords that are highly probable to appear in documents of interest. Besides, they have informed their expert knowledge of how the identified keywords are likely to appear in relevant documents. This provides the basis for a deep learning-based classification model trained with a small portion of labelled documents. Training a classification model, especially a deep learning-based classification model, with a small portion of documents will overfit the training data~\citep{bashar2018cnn,Bashar2020RegularisingSet,Bashar2021}.  But we conjecture that by integrating expert knowledge within a  classification model, we can reduce the number of labelled documents required for building an accurate model. We investigate how such prior knowledge can be integrated into the machine learning model for relevant document detection. Specifically, we investigate what kind of knowledge should be included, how this knowledge can be represented for a deep learning-based text classifier and wherein the machine learning pipeline this knowledge should be integrated. Our experimental results in Section~\ref{sec:results} show that by integrating Expert Informed Knowledge into the classification model can significantly improve the accuracy while requiring a small number of labelled documents.  

In sum, this paper makes four major contributions. 1) It proposes using machine-based deep learning as  a timely and effective means of greatly improving the accuracy and timeliness of research to assist First Nation communities in the repatriation and reburial of the remains of their ancestors. 
2) It proposes a new technique of document representation using context and expert-informed knowledge to improve the performance of deep learning based text classifiers.
3) It proposes a novel Informed Neural Network model for relevant document detection. By including expert informed knowledge within classification boundary decisions, the model requires a significantly less number of labelled documents for training without sacrificing accuracy.
4) It proposes a new centrality measure, ‘Keyword Centrality’, for mathematically modelling expert knowledge of how keywords are likely to appear in relevant documents.

\section{Related Work}
This section presents the related work in the areas of text classification and informed Machine Learning.

\subsection{Relevant Documents Detection: Text Classification}
Detecting relevant documents returned by a search engine  %to Indigenous Human Remains 
falls into the research area of text classification. Popular traditional text classification algorithms include Random Forest (RF) \citep{liaw2002classification}, k-Nearest Neighbours (kNN) \citep{weinberger2009distance}, Ridge Classifier (RC) \citep{hoerl1970ridge} and many others. Performance of these traditional machine learning algorithms depends on feature engineering and feature representation \citep{davidson2017automated,xiang2012detecting}. Besides, these algorithms are based on bag-of-words representation. The bag-of-words approach is straightforward and usually has a high recall, but it results in a higher number of false positives because the presence of general words and keywords causes these documents to be misclassified as relevant \citep{kwok2013locate}.

Detecting relevant documents, especially historical documents (e.g. relevant to Indigenous Human Remains) is a challenging text classification task because the words and language used in those documents are quite different from modern text documents. The problem is further challenged by the fact that most of the historical documents are OCRed for digitization that introduces a lot of noise (e.g. wrong character and word detection, failing to keep the right text structures, mixed-up multiple stories as OCR fails to separate adjacent stories in the paper face). 

Text classification research applications use syntactic features to identify the relevant documents. For instance, if someone was researching historical examples of antisemitism some relevant keywords would be \emph{kill} and \emph{Jews} (verb and noun occurrences) \citep{gitari2015lexicon} and the syntactic structures $<$intensity$>$$<$user intent$>$$<$hate target$>$ (e.g. \emph{I f$*$cking hate Jews people}) could be used for detecting hate speech~\citep{silva2016analyzing} in documents. Such feature engineering requires both linguistic and domain expertise that is expensive.  

Recently, neural network-based classifiers have become popular as they automatically learn abstract features from the given input feature representation \citep{badjatiya2017deep}. Input to these algorithms can be various forms of feature encoding, including many of those used in traditional methods. Algorithm design in this category focuses on the selection of the network topology to automatically extract useful abstract features. Popular network architectures are Convolutional Neural Network (CNN), Recurrent Neural Networks (RNN) and Long Short-Term Memory network (LSTM). CNN is well known for extracting patterns similar to phrases and $n$Grams  \citep{badjatiya2017deep}. On the other hand, RNN and LSTM are effective for sequence learning such as order information in text \citep{badjatiya2017deep}. The CNN model has been successfully used for sentence and noisy tweet classification \citep{kim2014convolutional,bashar2018cnn}.

The most recent success of neural network-based classifiers comes from transfer learning \citep{Bashar2020RegularisingSet,Bashar2021,Bashar2021ActiveTask,Liu2019RoBERTa:Approach,Devlin2018BERT:Understanding,Yang2019XLNet:Understanding}. Transfer learning is an approach in machine learning where knowledge gained in one model, such as through pretraining, is reused in another model \citep{Bashar2021,Bashar2020RegularisingSet}. Common pretraining in text data is conducted through language models. Popular pretrained models are RoBERTa \citep{Liu2019RoBERTa:Approach}, BERT \citep{Devlin2018BERT:Understanding} and XLNet \citep{Yang2019XLNet:Understanding}. However, experiments in Section \ref{sec:results} show that existing transfer learning-based models such as RoBERTa \citep{Liu2019RoBERTa:Approach} do not perform well on historical datasets. These results reveal that if there are not enough training data and the pretraining domain of transfer learning-based models is significantly different from the domain where the model will be applied, neural network and transfer learning-based models do not perform well. 

In humanities and social sciences, large collections of labelled data do not exist to effectively train a deep learning-based classification model. The number of documents initially returned by the search engine such as Trove \citep{WhatTrove} for the keyword set (as shown in Table \ref{tab:R3Keywords}) with OR operator is massive for two reasons: (a) relevant documents reside in the general span of any human remains while the researchers need documents of Indigenous human remains, e.g. news articles on any human remains collection, auction, selling, discovery, outcome scientific analysis of human remains. (b) It is difficult to select which sub-set of keywords should be used with AND operator in the search engine to perform specific searches. An incorrect or very specific combination can miss many important documents, while a generic combination can return a lot of documents including many non-relevant. 

To the best of our knowledge, there exists no work that fits the mentioned purpose of finding relevant documents using a text classifier. The existing literature of digital humanities mostly uses (a) Named Entity Extraction (NER) for identifying key information from a document collection, (b) Word Embedding or word vector for finding semantically similar words or words that appear together, and (c) Topic modelling for finding subjects of discussion.

\subsection{Informed Machine Learning}
% hard coded references are from \citep{vonRueden2019InformedSystems}
Although machine learning has achieved great success, it has limits when dealing with a small set of training data \citep{bashar2018cnn,Bashar2020RegularisingSet,Bashar2021}. Integration of prior knowledge such as that informed by experts into the training process can potentially address the problem of small training datasets \citep{vonRueden2019InformedSystems}. This integration process is known as informed machine learning and is becoming common in many applications \citep{vonRueden2019InformedSystems}. Informed machine learning investigates how to improve machine learning models by integrating prior knowledge into the learning process. It combines data- and knowledge-driven approaches. 

\begin{figure*}[htb]
		\centering
		\includegraphics[width=0.8\textwidth]{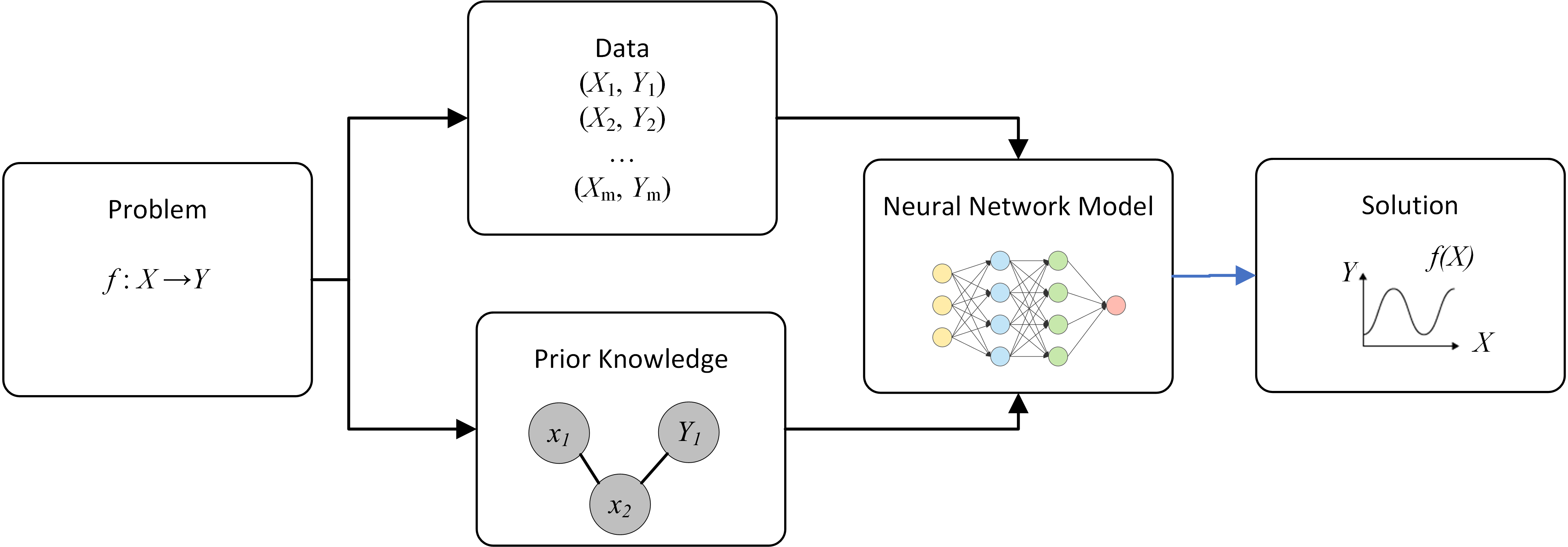}
		\caption{Informed Machine Learning Framework Adopted in this Research}
		\label{fig:IformedML}
\end{figure*}

A framework of Informed Machine Learning adopted in our work is shown in Figure \ref{fig:IformedML}. It enables experts to incorporate their domain knowledge into the machine learning process along with the datasets used for training the model. There is growing interest in informed machine learning in research areas such as engineering applications \citep{KarniadakisPhysics-informedLearning} and computer vision \citep{Marino2016TheClassification}. In these works, prior knowledge is used to define informative priors \citep{Heckerman1995LearningData} that regularise or constraint learning algorithms. For example, logic rules \citep{Diligenti2017IntegratingLearning,AKnowledge} and algebraic equations \citep{Daw2017Physics-guidedModeling,Stewart2016Label-FreeKnowledge} have been used to constraint loss functions. Knowledge graphs are used to enhance neural networks with information about relations between instances~\citep{Battaglia2016InteractionPhysics}, as well as to improve classification through CNNs by capturing relations between detected objects in computer vision \citep{Marino2016TheClassification}.
Knowledge graphs can be integrated into the learning process either explicitly or implicitly  \citep{Battaglia2016InteractionPhysics,Marino2016TheClassification,JiangHybridDetection}. Explicit systems use graph propagation and attention mechanisms, while implicit systems use graph neural networks with relational inductive bias \citep{vonRueden2019InformedSystems}. 

Informed machine learning for text data is mostly based on the principle of combinatorial generalisation that states that we construct new inferences, predictions, and behaviours from known building blocks \citep{Battaglia2018RelationalNetworks}. Combinatorial generalisation works by biasing the learning towards structured representations and computations, and in particular, systems that operate on graphs. Human cognitive mechanism represents a complex system with compositions of entities and their interactions \citep{Battaglia2018RelationalNetworks}. 
Humans draw analogies by aligning the relational structure between two domains. They can draw inferences on one domain based on the corresponding structural knowledge of another \citep{Hummel2003AGeneralization}. %That is, the world is understand in terms of composition \citep{Battaglia2018RelationalNetworks}. 
Combinatorial generalisation is recommended as a top priority for advancing modern AI \citep{Battaglia2018RelationalNetworks}. 

Inspired by the theory of combinatorial generalisation, we use four centrality measures, as described in Section \ref{sec:eik}, to represent the keywords and their word co-occurrence network based interactions in a given document (as shown in Fig. \ref{fig:IformedML}). Although integrating knowledge into machine learning for text classification is done through feature engineering or incorporating a sub graph (ontology) from an external source, there is not enough work on informed machine learning that utilise internal graph or network present in the text. Feature engineering requires manual efforts, and ontologies are mostly biased toward specific domains. To the best of our knowledge, this will be the first work to integrate an (internal) word co-occurrence network based on expert defined keywords into a deep learning model for text representation and classification.

\section{Relevant Document Detection: Problem Formulation}
%Assume a data analyst needs to collect relevant documents or texts to investigate an event of national interest (e.g., ). A set of keywords is identified assuming what might appear in those documents. However, those keywords can appear in other documents as well. As a result, a search engine returns a large number of documents where most of the documents are non-relevant for the investigation. They have checked a small portion of the documents and labelled them whether they are relevant for the investigation or not. Now, they need a classification model to detect rest of the documents returned by search engine whether they are relevant or not. For example, Return, Reconcile, Renew: Understanding The History, Effects & Opportunities Of Repatriation And Building An Evidence Base For The Future RRR team needs to collect documents that are relevant to Indigenous Human Remains. They want to analyse these documents for repatriation researchers to trace information about the location and provenance of Ancestral Remains held in museums and other collecting institutions. They have identify a set keywords as given in Table \ref{tab:R3Keywords}. They have used Trove to search relevant documents using these keywords but Trove has returned a large number of documents as these keywords appears in nonrelevant documents as well. They have labelled some of the returned documents as described in Section \ref{sec:r3trove}. They need a classification model that can detect relevant documents from the rest of the returned documents. 

Assume a historian needs to collect relevant documents to investigate an event of national interest. For example, researchers of the RRR network require to collect documents that are relevant to Indigenous Human Remains. This is done to trace information about the location and provenance of ancestral remains held in museums and other collecting institutions. This assists them in understanding the history, effects and opportunities of repatriation and building an evidence base for the future. A set of keywords (e.g. as given in Table \ref{tab:R3Keywords}) is identified assuming what might appear in those documents. However, those keywords can also appear in other documents. As a result, a search engine (such as Trove\footnote{A popular search engine service for browsing the digitised (archived) documents by the National Library of Australia.} \citep{WhatTrove}) returns a large number of documents where most of the documents are non-relevant to the investigation. With limited resources, only a small portion of the documents can be checked and labelled  whether they are relevant for the investigation or not. With the advances in text mining, a classification model trained on labelled data can be developed that can detect relevant documents within the rest of the returned documents. %Lets look 

Relevant document detection is a complex problem because relevance is determined by the implicit and explicit information of the domain that can be subtle and determined only through its context in the text and domain specific knowledge. Let $X$ be a text dataset that contains $N$ features or words and $C$ classes. Let $\mathbf{x} = \langle x_1,\dots x_n \rangle$ be a vector representing an instance in $X$. Let $K = \{k_1, k_2, \dots, k_n\}$ be a set of keywords. Let $Y$ be the set of $C$ classes. 

The relevant document detection is defined as a classification task that assigns an instance to a relevance class (or category) $Y_c$ based on the feature vector $\mathbf{x}$ and keywords $K$; i.e. $f \in \mathcal{F}: (X,K) \rightarrow Y$, where $f(\mathbf{x}, K) = \max_{Y_c} \, p(Y_c|\mathbf{x}, K)$. This ascertains that we need to know $p(Y_c|\mathbf{x}, K)$ for the relevance detection task. 
The joint probability $p(\mathbf{x}, Y_c, K)$ of $\mathbf{x}$, $Y_c$ and $K$ can be written as
\begin{equation}
    p(\mathbf{x}, Y_c, K) = p(Y_c|\mathbf{x}, K)p(\mathbf{x}, K)
\label{eq:bayes_simp_rearr}
\end{equation}
where $p(\mathbf{x}, K)$ is the prior probability distribution. The prior probability $p(\mathbf{x}, K)$ can be seen as a regulariser for $p(Y_c|\mathbf{x}, K)$ that can regularise modelling of the associated uncertainties of $p(\mathbf{x}, Y_c, K)$ \citep{Bashar2020RegularisingSet}. As $p(\mathbf{x}, K)$ does not depend on $Y_c$, this means that $p(\mathbf{x}, K)$ can be estimated independent of the class level $Y_c$. That is, $p(\mathbf{x}, K)$ can be estimated from prior knowledge of the interaction between $K$ and $\mathbf{x}$, e.g. interaction of keywords $K$ and $\mathbf{x}$ when they are represented in a graph or network. Estimating $p(\mathbf{x}, K)$ from prior knowledge and integrating it into a learning algorithm (shown as Equation \ref{eq:bayes_simp_rearr}) can be considered a form of Informed Machine Learning \citep{vonRueden2019InformedSystems}. A framework of Informed Machine Learning adopted in this research is given in Figure \ref{fig:IformedML}. The joint probability $p(\mathbf{x}, K)$ can be written as
\begin{equation}
    p(\mathbf{x}, K) = p(K|\mathbf{x})p(\mathbf{x})
    \label{eq:joint_prior}
\end{equation}
where $p(\mathbf{x})$ is the prior probability distribution of $\mathbf{x}$. $p(\mathbf{x})$ can be learned through transfer learning as described in our prior works \citep{Bashar2020RegularisingSet,Bashar2021,Bashar2021DeepDeployment} or it can be assumed as uniform distribution for simplicity. The term $p(K|\mathbf{x})$ can be interpreted as the probability of keyword set $K$ interacting in the text instance $\mathbf{x}$, while $p(K)$ is the probability of keyword set $K$ interacting in any text instance. Without considering the pattern of how $K$ appears in $\mathbf{x}$ versus other documents, $p(K|\mathbf{x})$ and $p(K)$ can be very similar as keywords can appear in both relevant and non-relevant documents. Hence, $p(K|\mathbf{x})$ may not bring useful information to improve Equation \ref{eq:joint_prior} and thereby will not improve the classification task in Equation \ref{eq:bayes_simp_rearr}. 

In this research, we propose and utilise four centrality measures to find patterns of the keyword set $K$ in each text instance $\mathbf{x}$ of the dataset $X$. We denote the patterns of keyword set $K$ in $\mathbf{x}$ as $\mathcal{K}$ and rewrite the Equations \ref{eq:joint_prior} and \ref{eq:bayes_simp_rearr} as Equations \ref{eq:joint_prior_pattern} and \ref{eq:bayes_simp_rearr_pattern} respectively.
\begin{equation}
    p(\mathbf{x}, \mathcal{K}) = p(\mathcal{K}|\mathbf{x})p(\mathbf{x})
    \label{eq:joint_prior_pattern}
\end{equation}
\begin{equation}
    p(\mathbf{x}, Y_c, \mathcal{K}) = p(Y_c|\mathbf{x}, \mathcal{K})p(\mathbf{x}, \mathcal{K}) = p(Y_c|\mathbf{x}, \mathcal{K})p(\mathcal{K}|\mathbf{x})p(\mathbf{x})
\label{eq:bayes_simp_rearr_pattern}
\end{equation}
We propose a model, namely Informed Neural Network (INN), %(described in Section \ref{sec:inn}, 
to integrate this Expert Informed Knowledge to the deep learning algorithm. We detail this model next. 

\section{Informed Neural Network Model for Relevant Document Detection}
\label{sec:inn}
%If there are not sufficient training data, especially when application domain is impacted by data shift, machine learning models can overfit and provide poor accuracy \citep{Bashar2021}. This problem can be potentially addressed by integrating prior knowledge into the training process \citep{vonRueden2019InformedSystems,Bashar2021}. This process of integration is known as Informed Machine Learning \citep{vonRueden2019InformedSystems}. 
We propose to improve the machine learning model by integrating prior or expert knowledge into the training process by combining data-and knowledge-driven approaches.  %is becoming popular in many research areas \citep{vonRueden2019InformedSystems}. 
 We propose an Informed Neural Network (INN) model for relevant document classification. The INN model uses a hybrid information source that consists of data and prior knowledge. The prior knowledge is pre-existent and independent of learning algorithms. The proposed INN model is shown in Figure \ref{fig:EIM_CNN}. It has two main parts: (a) Expert Informed Knowledge that constructs the \emph{Expert Informed Knowledge} matrix; and (b) Context text that constructs the \emph{Embedded Text} matrix. 

\subsection{Expert Informed Knowledge}
\label{sec:eik}
Expert knowledge can be considered to be intuitive knowledge that is held and implicitly validated by a particular group of experts \citep{vonRueden2019InformedSystems}. For a text collection, it is very difficult to source this knowledge manually as well as it poses the risk of subjectivity. In this paper, we use experts (i.e. information seekers) to identify a set of keywords that will be used to perform a search and collect the documents that may contain those keywords. We propose to explore the patterns of keywords that appear in each document and the collection, and use the patterns as expert knowledge. 

We aim to capture the expert knowledge of how the information seekers discern relevant hits based on their accumulated experience in understanding keywords, relationships between those keywords and the context of those keywords in the article. The following issues are indicative of the problems in capturing this information:
\begin{itemize}
    \item Many articles in \emph{digitized} (historical) newspapers appear as a composite of multiple articles (due to the ineffectiveness of OCR technologies in identifying borderlines). This means that keywords appear in the same article; however, many are unrelated articles and therefore do not signify a relevant hit.
	\item Not all keywords are equal. Some are more important than others. For example, an article with the keywords ‘grave, funeral, sepulcher, burial’ will more than likely not be relevant. What is needed is keywords across the different categories of terms for \emph{Identity}, \emph{Ancestral Remains}, \emph{Funerary Sites} and \emph{Mode of Acquisition}, as detailed in Table \ref{tab:R3Keywords}.
\item Keywords can be used to expand the number of results rather than limit the number of results. The historical relationship of keywords needs to be coded as expert knowledge. Within the listed categories of keywords in Table \ref{tab:R3Keywords}, there are various terms of description that were historically more common or used in a particular way. For example, ‘Aboriginal’ or ‘Australian’ was more common in scientific literature to describe Australian Indigenous remains than ‘nigger’ or ‘Black’ although not exclusively. The latter words are associated with pejorative articles and in the case of ‘Black’ could generate many irrelevant hits. The experienced human researcher can make this determination very quickly.  
\end{itemize} 

\begin{figure*}[htb]
		\centering
		\includegraphics[width=0.7\textwidth]{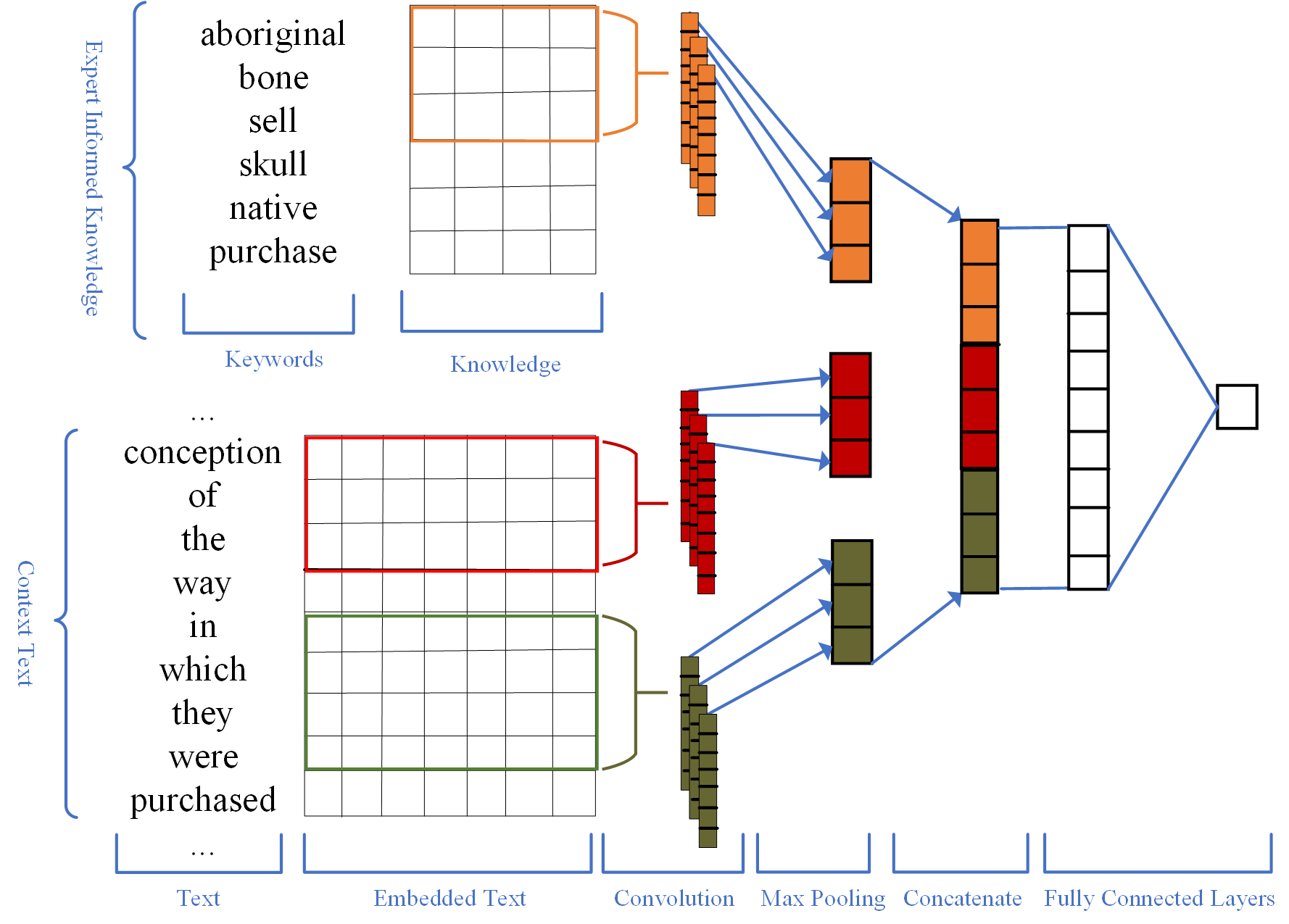}
		\caption{The Proposed Informed Neural Network Model Architecture}
		\label{fig:EIM_CNN}
\end{figure*}

As mentioned in Section \ref{sec:introduction}, RRR researchers informed their prior expert knowledge of how keywords are likely to appear in relevant documents (i.e. keywords appearing patterns). In this paper, we attempt to model this prior knowledge by utilising three centrality scores.  Centrality is a fundamental and popular topic in social network studies. Centrality ranks nodes within a network or graph according to their position in the network. Centrality is used in many applications such as identifying influential entities in a social network \citep{Bonacich2015PowerMeasures}, key nodes in the network \citep{Borgatti2005CentralityFlow}, super-spreaders of disease \citep{Bauer2012IdentifyingApproach} and Semantic Brand Score analysis \citep{FronzettiColladon2018}. %R3 experts in our team observed that if the keywords appear closely in a document, the document is relevant to R3 analysis. Based on this observation, 

\begin{figure*}[htb]
		\centering
		\includegraphics[width=0.7\textwidth]{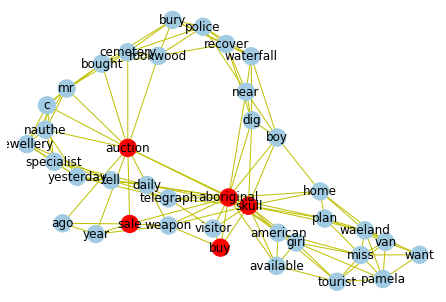}
		\caption{A word co-occurrence network generated from a document relevant to Indigenous human remains. Red nodes represent the keywords used in information seeking. Blue nodes represent the words present in the document.}
		\label{fig:doc_network}
\end{figure*}

We present each document as an undirected graph or network where each unique word is a node in the graph. Let us consider a word co-occurrence network \citep{Colladon2021LookNetworks} constructed from a document using a sliding window (i.e. words are assumed connected if they appear in the same sliding window). It is represented as an undirected graph $G = (V, E)$, where $V$ is the set of nodes corresponding to the unique words in the document and $E = (a, b): a,b \in V$ is the set of edges where each edge represents two connected words in the document. Figure \ref{fig:doc_network} shows an example of the undirected graph representing the word co-occurrence network of a document relevant to Indigenous human remains.

For a given keyword, we propose to use four centrality scores that reveal probabilistic interaction between a text and the patterns of a given keyword set. 
These four centrality scores are \emph{Keyword Centrality}, \emph{Betweenness Centrality}, \emph{Degree Centrality} and \emph{Prevalence}. We propose a Keyword Centrality score based on the spreading activation theory \citep{Collins1975} and estimate it as described below. Degree Centrality and Betweenness Centrality are two of the most famous centrality estimates \citep{FronzettiColladonid2020DistinctivenessNetworks}. The fourth score Prevalence is commonly used in relevance discovery \citep{Bashar2018InterpretationPatterns,alharbi2018random}.

\bigskip
\noindent \textbf{Keyword Centrality}: % similar to harmonic centrality but the motivation is Eigenvector centrality
\label{sec:Keyword_Centrality}
Keyword Centrality is proposed to estimate the relative interaction importance of keywords based on their co-occurrence patterns in a document. In other words, it can indicate how relevant a document is to a specific keyword $\hat{w}$ in a set of keywords $\hat{W}$. The centrality of a node in a network strongly depends on its neighbouring nodes. For example, the popular \emph{Page Rank} \citep{WhatWebmasters} measure calculates how `important' a web page is according to its connections, a web page connected with important pages receives a higher score. From a social network point of view, a few connections with important nodes can be enough to make a node important \citep{FronzettiColladonid2020DistinctivenessNetworks}. We argue that the highly connected nodes can be considered important, similar to the real world, where the social context collaboration between important persons makes them more influential. From the users' point of view, keywords are important indicators of the information that they are looking for in a document. Therefore, we consider keywords as important nodes in the co-occurrence network. We conjecture that a keyword connected to other keywords in the co-occurrence network should get higher centrality scores as compared to the keywords that are not.

We adopt the spreading activation theory \citep{Collins1975} to estimate Keyword Centrality. According to this theory, a semantic search in a network can be viewed as an activation spreading from two or more nodes corresponding to the keywords until an intersection is found. The activation spreads from a node by decreasing gradient \citep{Collins1975}. That is, the activation is similar to a signal from a source that is attenuated as it travels outward. 

\begin{figure*}[htb]
		\centering
		\includegraphics[width=0.6\textwidth]{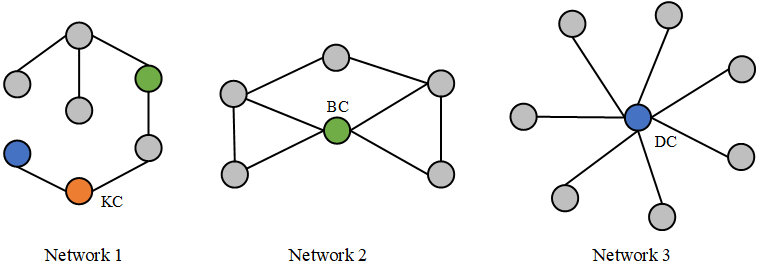}
		\caption{KC, BC and DC in three toy co-occurrence networks. Each node represents a word and each edge represents the co-occurrence of connected words. Blue, orange and green nodes are three keywords.}
		\label{fig:CentInNetwork}
\end{figure*}

Let us suppose that every keyword node spreads activation (i.e. we assume other words do not spread activation) along the co-occurrence network through its edges concurrently. The amount of activation reached to node $\hat{w}$ from node $\dot{w}$ per unit of distance is $\frac{a}{d(\hat{w}, \dot{w})}$, where $a$ is the activation initiated at $\dot{w}$ and $d(\hat{w}, \dot{w})$ is the shortest path distance between nodes $\hat{w}$ and $\dot{w}$ in the network. When there is no path in the network between $\hat{w}$ and $\dot{w}$, we assume $d(\hat{w}, \dot{w}) = \infty$, and consistently, the amount of spreading activation is 0. The Keyword Centrality of $\hat{w}$ in network $G$ is the total amount of activation reached to $\hat{w}$ over the network and its initial activation, which is estimated as follows.
%In a given set of keywords $\hat{W}$, the Keyword Centrality of a keyword $\hat{w}$ is estimated as the sum of inverse shortest path between $\hat{w}$ and other keywords as follow
\[
    KC(\hat{w}) = a + \sum_{\dot{w} \in \hat{W}-\hat{w}} \frac{a}{d(\hat{w}, \dot{w})}
\]
% and $d(\hat{w} \neq \dot{w})$.
This research uses initial activation $a = 1$. In this case, Keyword Centrality $KC(\hat{w})$ of a keyword $\hat{w}$ can be interpreted as its closeness (inverse of distance) to other keywords in the co-occurrence network, which is similar to proximity search \citep{Tao2007AnRetrieval}. Figure \ref{fig:CentInNetwork} shows an example of the calculation of KC.
In Network 1, the initial activation for each blue, orange and green node is 1. The amount of activation reached from the blue node to the orange node is $\frac{1}{1}$ and the amount of activation reached from the green node to the orange node is $\frac{1}{2}$. The total amount of activation reached to the orange node over the network is $\frac{1}{1}+\frac{1}{2}$. Therefore, KC for the orange node (keyword) in Network 1 is $1+\frac{1}{1}+\frac{1}{2} = 2.5$. Similarly, KC for the blue node and green node are $1 + \frac{1}{1} + \frac{1}{3} = 2.33$ and $1 + \frac{1}{2} + \frac{1}{3} = 1.83$ respectively. 

The higher the closeness, the more important the keyword is. 
For example, KC for the orange node (keyword) in Network 1 of Figure \ref{fig:CentInNetwork} is $1+\frac{1}{1}+\frac{1}{2} = 2.5$. The higher the closeness, the more important the keyword is. %Proximity search \citep{} uses distance between two or more separately matching term occurrences within a document, where distance is the number of intermediate words. Proximity search does not utilise the implicit connection between terms. For example, if A immediately occurs with B and after 100 words B immediately occurs with C. The distance between A and C according to proximity search is 102, while distance between A and C according to Keyword Centrality is just 2, as it estimates distance in the cooccurrence network. However, advanced proximity search is similar to our proposed keyword centrality. 

\bigskip

\noindent \textbf{Betweenness Centrality}: % aka Connectivity
Betweenness Centrality $BC(\hat{w})$ of keyword $\hat{w}$ estimates its connectivity with respect to a general discourse (i.e. which social context the keyword is representing in the document) \citep{FronzettiColladon2018}. $BC(\hat{w})$ represents the ability of keyword $\hat{w}$ to act as a bridge between nodes in the co-occurrence network \citep{Freeman1977ABetweenness} and show their dependence on the keyword $\hat{w}$. Connectivity is widely used in social network analysis as a measure of influence or control of information that goes beyond direct links. It is estimated as 
\[
BC(\hat{w}) = \sum_{j \neq k} \frac{s_{jk}(\hat{w})}{s_{jk}}
\]
where $s_{jk}$ is the number of the shortest paths linking any two nodes $j$ and $k$ in a co-occurrence network, and $s_{jk}(\hat{w})$ is the number of shortest paths that contain the keyword $\hat{w}$. For example, in Network 2 of Figure \ref{fig:CentInNetwork}, BC for the green node (keyword) is $\frac{0}{1}+\frac{1}{1}+\frac{1}{1}+\frac{1}{2}+\frac{0}{1}+\frac{1}{1}+\frac{1}{2}+\frac{1}{1}+\frac{1}{2}+\frac{1}{1}+\frac{1}{1}+\frac{1}{1}+\frac{0}{1}+\frac{0}{1}+\frac{1}{2} = 9$.

\bigskip
\noindent \textbf{Degree Centrality}: % aka Diversity
Degree Centrality $DC(\hat{w})$ of keyword $\hat{w}$ estimates the heterogeneity of words surrounding the keyword $\hat{w}$ \citep{FronzettiColladon2018}. The value of $DC(\hat{w})$ is high when $\hat{w}$ co-occurs with many different words. Whereas the value of $DC(\hat{w})$ is low when $\hat{w}$ co-occurs with a small set of words. Degree Centrality is estimated by counting the number of edges in a co-occurrence network directly connected to the keyword $\hat{w}$ \citep{Freeman1978CentralityClarification}.
\[
    DC(\hat{w}) = \sum_j a_j(\hat{w})
\]
where $a_j(\hat{w})$ is the adjacency between keyword $\hat{w}$ and node $j$ in the co-occurrence network. Value of $a_j(\hat{w})$ is 1 if $\hat{w}$ and $j$ are directly connected, otherwise 0. For example, in Network 3 of Figure \ref{fig:CentInNetwork}, DC score for the blue node (keyword) is 7 as there are seven nodes directed connected to the blue node. A high degree centrality score implies that a keyword co-occurs with a larger than average number of words in a document.

\bigskip
\noindent \textbf{Prevalence}:
Prevalence $PREV(\hat{w})$ of keyword $\hat{w}$ measures the number of times $\hat{w}$ occurs in a document \citep{FronzettiColladon2018}. Prevalence is associated with the idea of keyword awareness assuming that when the keyword occurs frequently, its recognition and recall are increased.  $PREV(\hat{w})$ of keyword $\hat{w}$ is calculated as follows.
\[
    PREV(\hat{w}) = f(\hat{w})
\]
where $f(\hat{w})$ is frequency of $\hat{w}$ in the document.%, $avg$ is average word frequency, $std$ is standard deviation of word frequency. 

\bigskip
\noindent \textbf{Expert Informed Knowledge Matrix}

\noindent A set of keywords (identified by domain experts or information seekers) will form the basis of an Expert Informed Knowledge matrix as shown in Figure \ref{fig:EIM_CNN}. Expert Informed knowledge $EIK(\hat{w})$ of keyword $\hat{w}$ is represented as a vector of four scores namely Keyword Centrality, Betweenness Centrality, Degree Centrality and Prevalence as follows. 
\[
    EIK(\hat{w}) = [ KC(\hat{w}), \; BC(\hat{w}), \; DC(\hat{w}), \; PREV(\hat{w}) ]
\]

The centrality measures, Betweenness Centrality, Degree Centrality and Prevalence, are known to be effective for analysing topical keyword impact and Brand impact \citep{bashar2022deep} \citep{FronzettiColladon2018}. Whereas, Keyword Centrality has similar properties of well-known proximity search \citep{Tao2007AnRetrieval} and Page Rank measure \citep{WhatWebmasters}. We conjecture that the discerning process of information seekers for deciding on relevant documents can be included in the machine learning model by capturing and representing the relative interaction importance of keywords for observing how keywords appeared in historical documents. 

\subsection{Context Text}
In addition to including Expert Informed Knowledge, it is critical that we add data-led information to the deep neural network model to learn the underlying data distribution. Such data context helps machine learning models to learn patterns that are not easy to model mathematically. Context text is a document instance consisting of a sequence of $m$ words or features, i.e. $\mathbf{x} = \langle x_1,\dots x_m \rangle$. Word embedding maps each word $x$ to a $n$ dimensional vector of real numbers. We use word embedding to represent each word $x \in \mathbf{x}$ in an $n$-dimensional word vector $w \in \mathbb{R}^{n}$. An instance $\mathbf{x}$ with $m$ words is represented as an Embedded Text matrix $\mathbf{x} \in \mathbb{R}^{m\times n}$ as shown in Figure \ref{fig:EIM_CNN}. Let us assume that we have a collection of such instances labelled as relevant or irrelevant by domain experts or information seekers. 

\subsection{Relevant Document Classification}
Knowledge-driven inputs (i.e the Expert Informed Knowledge matrix) and data-driven inputs (i.e. the Embedded Text matrix) are the inputs to the Convolutional Neural Network (CNN) model. We propose to use the CNN model to deal with the OCRed historical documents as CNN is known to be suitable for noisy text data that can detect location invariant patterns and sub patterns~\citep{bashar2018cnn}. Besides, the way convolution filters operate is analogous to searching in the local proximity of a network, which makes the CNN model an ideal model for combining Expert Informed knowledge with context knowledge.

The convolution operation is applied to both the Expert Informed Knowledge matrix and Embedded Text matrix with one stride. Applying convolution on the Expert Informed Knowledge matrix facilitates the finding of patterns and interactions among the centrality scores. This allows emphasising the importance of keywords in the document considering all measures where each of them is measured in isolation. Convolution applied on the Embedded Text matrix is supposed to find patterns analogous to $n$Grams emphasising patterns in the data-driven inputs~\citep{bashar2018cnn}. Each convolution operation applies a filter $\mathbf{F}_i \in \mathbb{R}^{h \times n}$ of size $h$, where $n$ is the dimension of embedding or word vector. Empirically, based on the accuracy improvement in ten-fold cross-validation experiments, we used 128 filters for $h = 3$, 256 filters for $h = 4$ and 512 filters for $h = 5$ on the Embedded Text matrix. We used 512 filters for $h =3$ on the Expert Informed Knowledge matrix.

Convolution is a function $\mathcal{C}(\mathbf{F}_i, \mathbf{M}) = R(\mathbf{F}_i \cdot \mathbf{M}_{k:k+h-1})$, where $\mathbf{M}_{k:k+h-1}$ is the $k$th vertical slice of a matrix from position $k$ to $k+h-1$, $\mathbf{F}_i$ is the given filter and $R$ is a ReLU function. Function $\mathcal{C}(\mathbf{F}_i, \mathbf{M})$ produces a feature $c_k$ for each slice $k$, resulting in $m-h+1$ features. 

We apply the max-pooling operation over these features and take the maximum value, i.e. $ \hat{\mathcal{C}}_i = \max \mathcal{C}(\mathbf{F}_i, \mathbf{M})$. Max-pooling is carried out to capture the most important feature of each filter. As there are a total of 1408 filters ((128+256+512)+512) in the proposed model, the 1408 most important features are learned from the convolution layer. 
	
These features are passed to a fully connected hidden layer with 256 perceptrons that use the ReLU activation function. This fully connected hidden layer allows learning the complex non-linear interactions between the features from the convolution layer and generates the 256 higher-level new features to learn to distinguish between relevant and non-relevant documents. Finally, these 256 higher-level features are passed to the output layer with a single perceptron that uses the sigmoid activation function. The perceptron in this layer generates the probability of a document being relevant. 
	
We randomly dropout a proportion of units from each layer except the output layer by setting them to zero. This is done to prevent co-adaptation of units in a layer and to reduce overfitting \citep{Hinton2012ImprovingDetectors}. We empirically dropout 50\% units from the input layer, the filters of size 3 and the fully connected hidden layer. We dropout only 20\% units from the filters of sizes 4 and 5 and the filter used on the Expert Informed Knowledge matrix.

\section{Empirical Evaluation}
Extensive experiments are conducted to evaluate the accuracy of the proposed method for relevant document detection. We used six standard classification evaluation measures \citep{Bashar2020RegularisingSet}: Accuracy, Precision, Recall, F$_1$ Score, Cohen Kappa Score (CK Score) and Area Under Curve (AUC). A description of evaluation measures is given in Appendix A.

\subsection{Data Collection}
We used two data collections R3Trove and Reuters Corpus Volume I (RCV1) in our experiments. R3Trove has one topic and RCV1 has 50 topics used in the experiments. %Each topic simulates a single user who might be seeking information for analysis.
Each topic can be seen as an information-seeking need of practitioners/researchers for conducting analyses on the retrieved documents based on the search terms contained within the topic. The following subsections give a brief description of these datasets. 

\subsubsection{R3Trove}
\label{sec:r3trove}
National Library of Australia (NLA) has collected and digitised relevant historical documents, images, and other cultural artefacts on a large-scale \citep{Kutty2020PaperMineraArticles}. NLA contains over 6 billion digital items on various topics by aggregating its own and other digital collections in partnership with other Australian state libraries, museums art galleries, media, government and community organisations \citep{WhatTrove}. The goal is to advance public knowledge of history and heritage by providing free access to these collections to users who are interested in curating particular aspects of Australia's history, heritage, and culture. NLA has developed a faceted search engine named Trove \citep{WhatTrove} to discover these items of interest. 

The most remarkable collection indexed in Trove is the Australian Newspaper Service. By using optical character recognition (OCR), NLA has digitised Australia's surviving newspapers starting from the first years of colonisation \citep{Kutty2020PaperMineraArticles}. 
%This digitisation provides the exciting potential of using advanced data mining techniques, entity recognition, relationship extraction and geotagging to identify and analyse relations between historical people, places, concepts, and many other entities in past time and space.
We collected news articles relevant to Indigenous Human Remains from Trove in three iterations using the following four steps.
\begin{itemize}
    \item STEP 1: Using the keywords in Table \ref{tab:R3Keywords} with OR operator, collect news articles from Trove utilising its associated Application Programming Interface (API).
    \item STEP 2: Due to the generality of keywords, STEP 1 can return lots of news articles not relevant to Indigenous Human Remains. It is not possible to manually go through each document and assess them. We use an LSTM-based classification model, described in Section \ref{sec:baselines}, to filter out non-relevant documents. 
    \item STEP 3: Domain experts manually label the remaining (and manageable) news articles whether they are relevant to Indigenous Human Remains. Then the labels are cross checked by a coordinator for correctness.
    \item STEP 4: Retrain the LSTM model using the labelled news articles collected so far. Go to STEP 2 for the next iteration.
    
\end{itemize}
The collected labelled news articles, following this multi-step process, are called R3Trove. Both news title and news text are used as the content of an article, and each article constitutes a document. R3Trove has a total of 1432 documents of which 844 documents are labelled relevant and 588 documents are labelled non-relevant. Out of the R3Trove collection, 90\% is used for training and validation of the proposed INN model and 10\% is used for testing.

The LSTM is used in this labelling process due to its flexibility in accepting variable-length documents, given that documents in R3Trove come in various lengths. The use of an LSTM model (but not a CNN model) also ensures that the labelling process does not introduce any bias and results in favouring the proposed INN model that uses a CNN component.

% Table generated by Excel2LaTeX from sheet 'Sheet4'
% \begin{table}[htbp]
%   \centering
%   \caption{Keywords Used in R3Trove Collection}
%     \begin{tabular}{p{30em}}
%     \hline
%     aborigin, native, indigenous, skull, crania, skeleton, mummy, bone, hair, tattoo, purchase, sell, sold, cost, price\\
%     \hline
%     \end{tabular}%
%   \label{tab:R3Keywords}%
% \end{table}%

% Table generated by Excel2LaTeX from sheet 'Sheet2'
\begin{table*}[htbp]
  \centering
  \caption{Keywords Used in R3Trove Collection}
    \begin{tabular}{l|p{30em}}
    \hline
    Feature Group & Associated Keywords \\
    \hline
    Identity & aborig*, native, black, negro, nigger, Indigenous, dusky, [maori, indian, moriori], torres strait \\
    \hline
    Ancestral Remains & skull, crania, bones, skeleton, remains, teeth, mummy, head, pickled, hair, cast, replica, modified, decorated, preserved heads, baked heads, moko, tattooed. \\
    \hline
    Funerary Sites & grave, funeral, sepulchre, burial, funer*, coffin, platform, massacre, internment \\
    \hline
    Mode of Acquisition & cost, sale, sold, sell*, purchase*, exchange*, auction, price, dollar, pound, deal*, shilling, (currency symbols \$,£, /-); buy, bought, value, trade, commodity; profit \\
    \hline
    \end{tabular}%
  \label{tab:R3Keywords}%
\end{table*}%

\subsubsection{Reuters Corpus Volume I}
Reuters Corpus Volume I (RCV1) is a standard data collection from the TREC-10/2001 filtering track \citep{robertson2002trec} provided by Reuters Ltd. It has English language news stories that cover a large spectrum of topics and contemporary information written by journalists.
%, RCV1 consists of 806,791 news stories. The training set contains 23,307 documents, and the testing set contains 783,484 documents. 
Both the story title and text are used as the content of a story, and each story constitutes a document.

RCV1 has 100 topics. Each topic has a corresponding set of documents and a manual specification of information needs written by linguists. The documents in the first 50 topics are labelled by domain experts as either relevant to the topic specification or non-relevant. For each topic, domain experts divided the documents into a training set and a testing set. Buckley and Voorhees \citep{buckley2000evaluating} showed that the first 50 topics are stable and sufficient for maintaining the accuracy of the evaluation measures. Therefore, the first 50 topics are used in this research. The set of words in each topic name (e.g. Child Custody Cases) found in the topic specification is used as the keywords set of the topic. There are a total of 2,704 documents in the training set with a minimum of 13 and a maximum of 198 documents in each topic. On the other hand, there is a total of 18,901 documents in the test set with a minimum of 199 and a maximum of 597 documents in each topic.

% \subsection{Evaluation Measures}
% We used six standard classification evaluation measures: Accuracy, Precision, Recall, F$_1$ Score, Cohen Kappa Score (CK Score) and Area Under Curve (AUC). A description of evaluation measures is given in Appendix A.
\subsection{Baseline Models}
\label{sec:baselines}
We have implemented 8 baseline models to compare the performance of the proposed INN model. For all neural network-based models, hyperparameters are manually tuned based on cross-validation.

\begin{itemize}%[leftmargin=*]

    \item Feedforward Deep Neural Network (DNN) \citep{glorot2010understanding}: It has five hidden layers, each layer containing eighty units, 50\% dropout applied to the input layer and the first two hidden layers, softmax activation and 0.04 learning rate. To keep this model basic, term frequency vector of Text is fed as input to this model. 

    \item Long Short Term Memory (LSTM)  \citep{hochreiter1997long}: Input to this model is Embedded Text Matrix. It has 100 units, 50\% dropout, binary cross-entropy loss function, Adam optimiser and sigmoid activation.

    \item Convolutional Neural Network (CNN): This model uses only the Embedded Text Matrix part of INN. In this CNN model, the first layer is a convolution layer with 1024 filters followed by Max Pooling, the second layer is a fully connected layer with 256 perceptrons followed by ReLU activation function, the final layer is a classification layer with single perceptron followed by Sigmoid activation function. The rest of the hyperparameters of this CNN model is set as in \citep{bashar2018cnn}.
    
    \item RoBERTa \citep{Liu2019RoBERTa:Approach}: Input to this model is Embedded Text Matrix, but it uses the embedding defined by the state-of-the-art language model RoBERTa~\citep{Liu2019RoBERTa:Approach}. RoBERTa is a retrained BERT with improved training methodology, more data and compute power. RoBERTa removes the \emph{Next Sentence Prediction} task from BERT's pre-training and introduces dynamic masking so that the masked token changes during each training epoch. RoBERTa used a total of 160 GB text data for pretraining in comparison to 13 GB text data used for training BERT$_{Large}$ and 126 GB text data for building XLNet$_{Large}$. RoBERTa used 1024 V100 Tesla GPUs running for a day during pretraining. As a result, RoBERTa is known to outperform both BERT$_{Large}$ \citep{Devlin2018BERT:Understanding} and XLNet$_{Large}$ \citep{Yang2019XLNet:Understanding}. 
    %\item Keyword Frequency Model (KFM):
    
    \item Expert Informed Knowledge Only Model (IKOM): This model is a variant of the proposed INN model. It uses only the Expert Informed Knowledge part of INN model and excludes the Embedded Text part as shown in Figure \ref{fig:EIM_CNN}. In other words, it does not take the Text input, it takes only keywords as input. 

    \item Non NN models including Random Forest (RF) \citep{liaw2002classification}, k-Nearest Neighbours (kNN) \citep{weinberger2009distance} and Ridge Classifier (RC) \citep{hoerl1970ridge}. Hyperparameters of all these models are automatically tuned using ten-fold cross-validation and GridSearch using sklearn library. The term frequency vector of Text is fed as input to these models.
\end{itemize}

\subsection{Experimental Results}
\label{sec:results}

\subsubsection{Results on R3Trove}
\label{sec:ResultsR3Trove}
Experimental results in Table \ref{tab:TroveResults} show that the proposed INN model performs better than all the baseline models. It achieves the best Accuracy (0.924), Precision (0.910), $F_1$ Score (0.943), CK Score (0.828) and AUC (0.901). The best Recall is achieved by the DNN model. However, its Precision and $F_1$ Score are the lowest of all models. When investigated, we observed that DNN predicts all documents as relevant, while many of them are not relevant, reflected by a poor AUC value. This is further emphasised by the CK Score of zero obtained from DNN, which means there are no agreements between ground truth and the prediction made by DNN.  

The second-best performance is obtained by RF. Three models RoBERTa, CNN and RC achieve similar performance and they provide the third-best performance. Even though RoBERTa is well known for its performance in text classification, its performance on R3Trove is not the best. The R3TRove dataset represents old historical documents that have a very different distribution than the data used to pre-train RoBERTa for transfer learning. Besides, RoBERTa is a very large neural network model. Overfitting of large neural network models, when trained with a small dataset, is usually compensated by their transfer learning capacity. The inadequacy of transfer learning of RoBERTa in R3Trove can be attributed to its poor performance. 

It is interesting to note that IKOM (knowledge information only) does not provide satisfactory performance as it lacks the context information based on the data distribution required for the training of the neural network. Similarly, CNN (context information only) does not provide a good performance due to the small training set. However, the proposed INN model (combining both knowledge and context information) achieves the best performance. This confirms the idea of informed machine learning, which combines prior knowledge with machine learning, can address small data problems for machine learning and lack of context problems for prior knowledge.

\begin{table*}[htbp]
  \centering
  \caption{Experimental Results of INN and baseline models on the R3Trove data}
    \begin{tabular}{l|cccccc|c}
    \hline
          & \multicolumn{1}{l}{Accuracy} & \multicolumn{1}{l}{Precision} & \multicolumn{1}{l}{Recall} & \multicolumn{1}{l}{F1 Score} & \multicolumn{1}{l}{CK Score} & \multicolumn{1}{l}{AUC} & \multicolumn{1}{l}{Mean Score} \\
    \hline
    INN & \textbf{0.924} & \textbf{0.910} & 0.978 & \textbf{0.943} & \textbf{0.828} & \textbf{0.901} & \textbf{0.914}\\
    IKOM   & 0.792 & 0.812 & 0.882 & 0.845 & 0.528 & 0.755 & 0.769 \\
    CNN   & 0.868 & 0.870 & 0.935 & 0.902 & 0.702 & 0.840 & 0.853\\
    LSTM  & 0.819 & 0.838 & 0.892 & 0.865 & 0.595 & 0.789 & 0.800\\
    DNN   & 0.646 & 0.646 & \textbf{1.000} & 0.785 & 0.000 & 0.500 & 0.596\\
    RoBERTa & 0.875 & 0.887 & 0.925 & 0.905 & 0.722 & 0.855 & 0.861\\
    kNN   & 0.861 & 0.869 & 0.925 & 0.896 & 0.688 & 0.835 & 0.846\\
    RC    & 0.875 & 0.887 & 0.925 & 0.905 & 0.722 & 0.855 & 0.861\\
    RF    & 0.889 & 0.889 & 0.946 & 0.917 & 0.750 & 0.865 & 0.876\\
    
    \hline
    \end{tabular}%
  \label{tab:TroveResults}%
\end{table*}%

\bigskip

\noindent \textbf{Ablation study}: We conducted an ablation study to see the importance of each of the centrality measures in making the Expert Informed Knowledge for machine learning. Results of this ablation study in Table \ref{tab:AblationSRSTrove} show how important each of the centrality measures is in contributing to Expert Informed Knowledge. %Outcome of this study is given in Table \ref{tab:AblationSRSTrove}. %Another ablation study is conducted to see how much each of the centrality measure can contribute to the relevant document selection problem, and the outcome is given in Table \ref{tab:AblationINNTrove}.Results in Table \ref{tab:AblationSRSTrove} show the significance of the proposed Keyword Centrality measure. 
As described in Section \ref{sec:baselines}, IKOM uses all four centrality measures KC, BC, DC and PREV. NPREV model assigns 0 to PREV value in IKOM, NDC assigns 0 to DC, NBC assigns 0 to BC and NKC assigns 0 to KC. Results in Table \ref{tab:AblationSRSTrove} show that IKOM is marginally affected by not including PREV and BC measures, but the performance is significantly affected by not including KC and DC. These results also show that each of the centrality measures can improve performance and the best performance is achieved when all four centrality measures are used together in INN as shown in Table \ref{tab:TroveResults}.

% Table generated by Excel2LaTeX from sheet 'Compare'
\begin{table*}[htbp]
  \centering
  \caption{Ablation study showing the significance of each centrality measure as Expert Informed Knowledge used in IKOM: R3Trove Dataset}
    \begin{tabular}{lcccccc}
    \hline
          & \multicolumn{1}{l}{Accuracy} & \multicolumn{1}{l}{Precision} & \multicolumn{1}{l}{Recall} & \multicolumn{1}{l}{F1 Score} & \multicolumn{1}{l}{CK Score} & \multicolumn{1}{l}{AUC} \\
    \hline
    IKOM   & 0.792 & 0.812 & 0.882 & 0.845 & 0.528 & 0.755 \\
    NPREV & 0.778 & 0.796 & 0.882 & 0.837 & 0.492 & 0.735 \\
    NBC & 0.750 & 0.766 & 0.882 & 0.820 & 0.417 & 0.696 \\
    NDC  & 0.729 & 0.793 & 0.785 & 0.789 & 0.411 & 0.706 \\
    NKC   & 0.688 & 0.735 & 0.806 & 0.769 & 0.289 & 0.639 \\
    \hline
    \end{tabular}%
  \label{tab:AblationSRSTrove}%
\end{table*}%

\begin{figure}[ht]
    \centering
    \includegraphics[width=.5\textwidth]{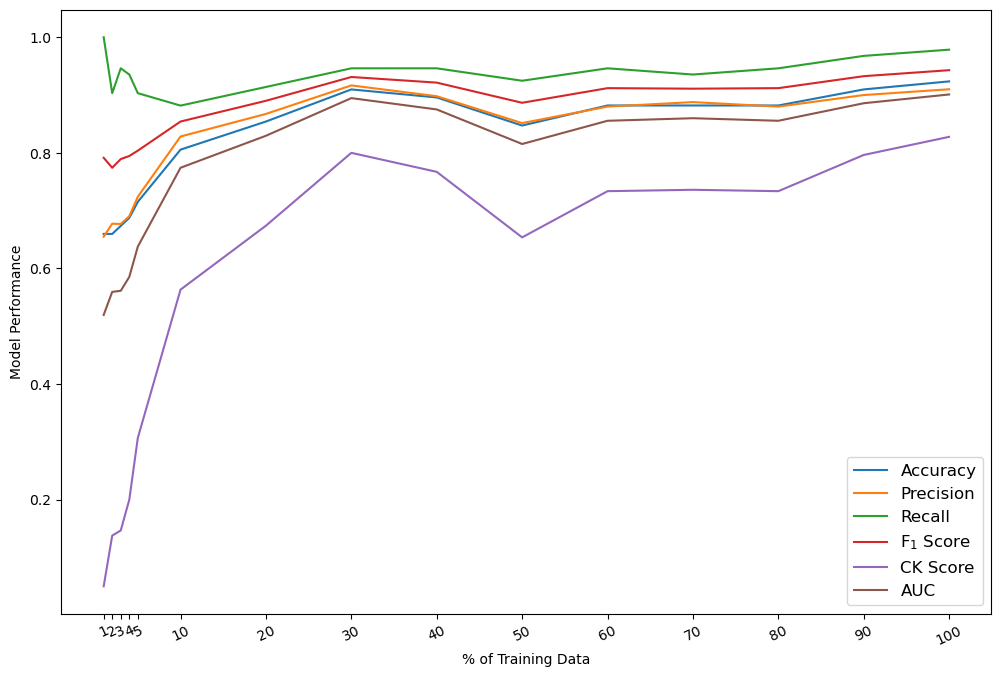}
    \caption{Impact of Training Data Percentage Change in the R3Trove Dataset to the INN model.}
    \label{fig:TrainingPercentChange}
\end{figure}

\noindent \textbf{Training Data Size}: We also conducted experiments to see the impact of training dataset size. The experimental result in Figure \ref{fig:TrainingPercentChange} shows that INN can provide reasonable performance with a very small amount of data. The model can work with as low as 1\% (12 documents) of the training data. With this amount of data, the model gives us Accuracy 0.660, Precision 0.655, Recall 1.00, F$_1$ Score 0.79, CK Score 0.05 and AUC 0.520. At this point CK score is low but with only 10\% (128 documents) of data CK score reaches to 0.56 and other measures improve significantly. With 30\% (386 documents) the model provides significantly high performance. At this point, it reaches Accuracy 0.910, Precision 0.917, Recall 0.946, F1 Measure 0.931, CK Score 0.800, AUC 0.895. After this, there is some fluctuation, but starting from 80\% (1030 documents) data, the performance steadily increases and provides maximum performance with 100\% (1288 documents) data. At this point, we get Accuracy 0.924, Precision 0.910, Recall 0.979, F$_1$ Score 0.943, CK Score 0.828, AUC 0.901.

\subsubsection{Results on RCV1}
\label{sec:ResultsRCV1}
\begin{figure*}[htb!]
    \centering
    \subfloat{{\includegraphics[width=3cm]{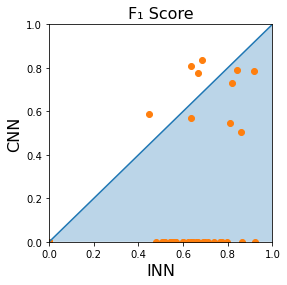}}}
    \quad
    \subfloat{{\includegraphics[width=3cm]{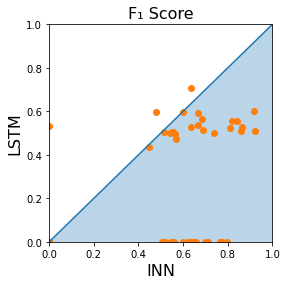}}}
    \quad
    \subfloat{{\includegraphics[width=3cm]{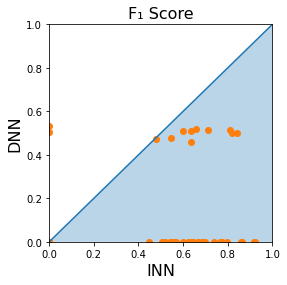}}}
    \quad
    \subfloat{{\includegraphics[width=3cm]{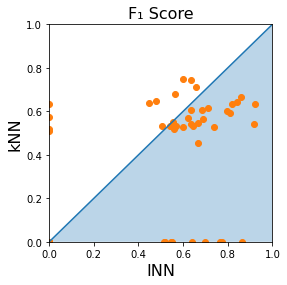}}}
    
    \subfloat{{\includegraphics[width=3cm]{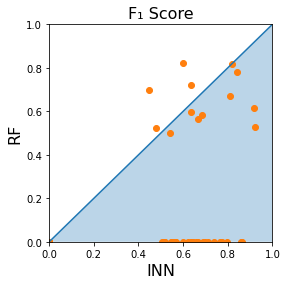}}}
    \quad
    \subfloat{{\includegraphics[width=3cm]{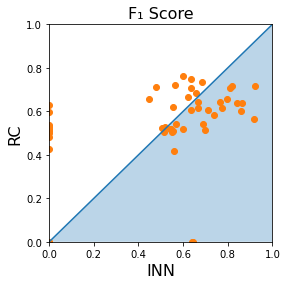}}}
    \quad
    \subfloat{{\includegraphics[width=3cm]{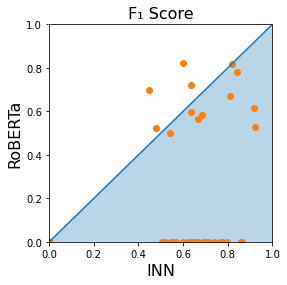}}}
    \quad
    \subfloat{{\includegraphics[width=3cm]{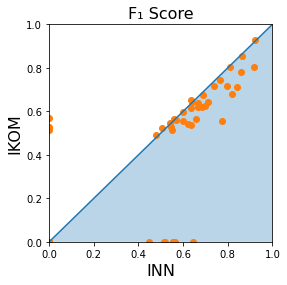}}}

    \caption{Pairwise performance comparison of INN against baseline models. Each dot indicates a topic in RCV1. A dot in the blue shade triangle indicates INN performs better than its competitor and dots in the white triangle indicate otherwise, with dots on the diagonal indicating the equal performance of both models.}%
    \label{fig:pwc}%
    \vspace{-.6cm}
\end{figure*}

Figure \ref{fig:pwc} shows the pairwise comparison of INN and each baseline model over the 50 topics. The number of dots in the blue shade triangle indicates the number of topics where INN performs better than a baseline model for the most important measure of classification namely F$_1$ measure. Results show that INN outperforms baseline models in the majority of topics as indicated by more dots in the blue shade triangle. 
%It is interesting to note that traditional machine learning such as RC, KNN and RF produced better results than deep neural models such as CNN, LSTM and DNN. Although the performance of RoBERTa is not so good comparatively traditional classifiers and INN, its performance is much better than these word embedding based models on this data. Since the nature of data used in training the language model in RoBERTa is close to the nature of RCV1, the performance is enhanced by using the language features. 
The well-known RoBERTa model did not also perform well in this dataset. Besides the small number of documents, this might be due to the document length of RCV1 dataset. RoBERTa can only consider 500 words in a document, the rest of the words in a document are ignored when fed into the model. The maximum document lengths in RCV1 are 7960 and 8119 for the training and testing sets respectively. Average document lengths in RCV1 are 421 and 520 for training and testing set respectively. It is interesting to note that on this dataset, non neural network models RC, RF and kNN perform significantly better than the neural network models except INN. This might be because neural network models are overfitting as many topics in RCV1 have a very small number of training documents, a minimum of 13 documents. We observed that for some topics, neural network models other than INN failed to detect any relevant documents. By introducing bias through the utilisation of L2 norm regularisation RC reduces overfitting and provides the second-best results. By introducing bias through the utilisation of Expert Informed Knowledge, INN can detect relevant documents even when the number of training documents is very small and provides the best results. When the number of documents increases, the detection performance improves.

%is found to be the second best performing model.  

\begin{figure*}[htb!]
    \centering

    \subfloat{{\includegraphics[width=4cm]{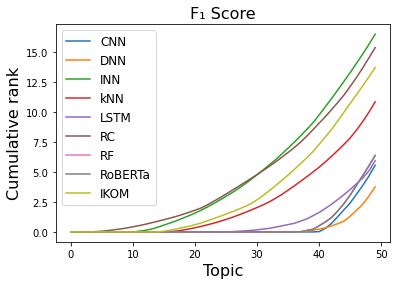}}}
    \quad
    \subfloat{{\includegraphics[width=4cm]{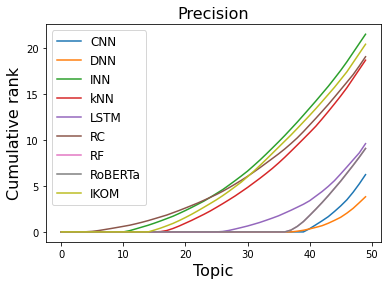}}}
    \quad
    \subfloat{{\includegraphics[width=4cm]{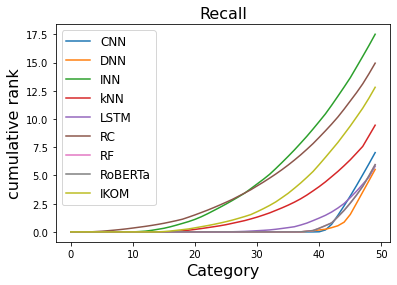}}}
    
    \subfloat{{\includegraphics[width=4cm]{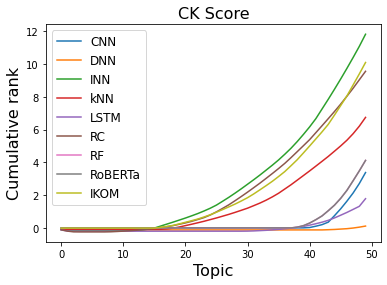}}}
    \quad
    \subfloat{{\includegraphics[width=4cm]{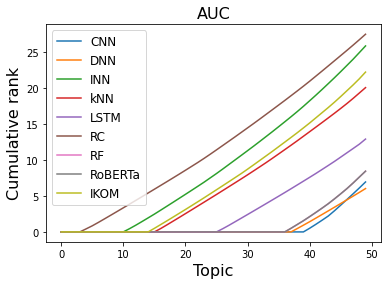}}}
    \quad
    \subfloat{{\includegraphics[width=4cm]{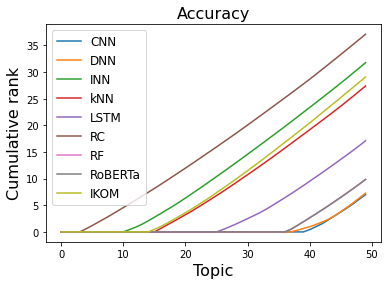}}}
    
    \caption{The cumulative ranking obtained by all methods by individually ranking each model per topic per measure}%
    \label{fig:cr}%
\end{figure*}

Figure \ref{fig:cr} shows the relative ranking of INN with baseline models by computing the cumulative ranking for all the models over the 50 topics. It is a method-wise accumulation of results measured on an evaluation criterion spanned across all topics. To compute the cumulative ranking, first, we rank the performance of each individual method for each topic based on a criterion and then sum up their ranks. Figure \ref{fig:cr} shows that INN ranks higher than all baseline models for F$_1$ Score, Precision, Recall and CK Score. INN achieves the second-best Accuracy and AUC. The best Accuracy and AUC are achieved by RC. However, RC has a significantly lower Recall value when compared with INN. This means RC misses many relevant documents. Also, Precision, F$_1$ Score and CK Score of RC are lower than those of INN. A similar pattern is observed in average performance shown in Table \ref{tab:RCV1Results}. 

Table \ref{tab:RCV1Results} shows the average performance of all methods on 50 topics of the RCV1 dataset. Results show that the proposed model INN performs the best. It gives the best Precision (0.430), Recall (0.350) and CK Score (0.236). INN achieves the second-best Accuracy (0.635) and AUC (0.516). The best Accuracy (0.742) and AUC (0.548) are achieved by RC. However, its Recall (0.299) value is significantly lower than that of INN. This means RC misses many relevant documents to identify. Also, the Precision (0.384), F$_1$ Score (0.308) and CK Score (0.191) of RC are lower than INN.  

\begin{table*}[htbp]
  \centering
  \caption{RCV1 Results (Average on 50 Topics.)}
    \begin{tabular}{l|cccccc|c}
    \hline
          & \multicolumn{1}{l}{Accuracy} & \multicolumn{1}{l}{Precision} & \multicolumn{1}{l}{Recall} & \multicolumn{1}{l}{F1 Score} & \multicolumn{1}{l}{CK Score} & \multicolumn{1}{l}{AUC} & \multicolumn{1}{l}{Mean Score} \\
    \hline
    INN & 0.635 & \textbf{0.430} & \textbf{0.350} & \textbf{0.330} & \textbf{0.236} & 0.516 & \textbf{0.416}\\
    IKOM   & 0.581 & 0.409 & 0.256 & 0.274 & 0.202 & 0.443 & 0.361\\
    CNN   & 0.140 & 0.125 & 0.141 & 0.111 & 0.067 & 0.139 & 0.120\\
    LSTM  & 0.342 & 0.192 & 0.116 & 0.119 & 0.036 & 0.258 & 0.177\\
    DNN   & 0.146 & 0.077 & 0.111 & 0.075 & 0.002 & 0.120 & 0.088\\
    RoBERTa & 0.272 & 0.195 & 0.159 & 0.153 & 0.087 & 0.225 & 0.182\\
    kNN   & 0.548 & 0.374 & 0.189 & 0.217 & 0.135 & 0.400 & 0.311\\
    RC    & \textbf{0.742} & 0.381 & 0.299 & 0.308 & 0.191 & \textbf{0.548} & 0.412\\
    RF    & 0.197 & 0.182 & 0.119 & 0.128 & 0.082 & 0.168 & 0.146\\
    \hline
    \end{tabular}%
  \label{tab:RCV1Results}%
\end{table*}%

\begin{figure}[htb!]
    \centering
    \includegraphics[width=8cm]{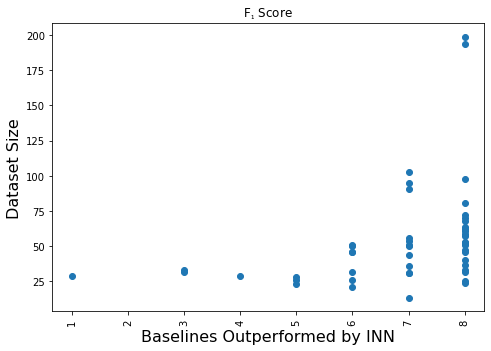}
    \caption{Each point in the scatter plot represents one of the 50 topics/datasets in the RCV1 collection. For a given dot, the horizontal axis shows the number of baseline models outperformed by INN (i.e. 1 indicates that INN outperforms only 1 baseline whereas 8 indicates INN performs the best) for this topic. There are a total of eight baseline models. The vertical axis shows the number of training documents available for this topic.}
    \label{fig:beat_dist}
    \vspace{-.6cm}
\end{figure}

Figure \ref{fig:beat_dist} shows an analysis of the INN model performance against the dataset size (i.e. the number of training documents per topic). This figure reveals that INN always performs better than at least one of the compared algorithms on any of the 50 topics. It has performed the best on 24 topics (i.e. outperformed all the 8 baseline models). The presence of the majority of dots (43 topics) appearing on the right side (6 or more baseline outperformed by INN) indicates INN outperforms the majority of the baseline models. Additionally, with the majority of the points (46) appearing on the upper part (trained with 25 or more documents) of this figure indicates that the performance of INN is better on relatively larger datasets than the baseline models. All these results indicate that Expert Informed Knowledge can generalise the machine learning model while training data can provide the context and specificity for learning. By combining data- and knowledge-driven approaches we can achieve the best of both worlds.

\section{Conclusion}
This paper proposes the use of deep learning, specifically an Informed Neural Network model, to enhance investigations into the theft, trade, and exchange of ancestral bodily remains of Australian and other First Nations peoples. The model employs expert informed knowledge and centrality measures to learn the distribution patterns in the data collection and is able to identify relevant documents with high accuracy, using significantly fewer labelled documents for training.

The proposed INN model was tested using both data provided by RRR researchers and a publicly available generic dataset. The results of the experiments showed that the INN model is generalizable to other datasets, despite being designed specifically for detecting relevant documents related to Indigenous Human Remains. In terms of performance, the INN model outperformed baseline models on both datasets. However, on the RCV1 dataset for a small number of topics, other models demonstrated slightly better performance than INN, particularly when the number of training documents was small. In the future, this research will further investigate these topics and modify the INN model accordingly. In addition, this study used only centrality measures as Expert Informed Knowledge. It may be useful to also investigate other types of prior knowledge, such as knowledge graphs, in order to improve the INN model. The keyword centrality measure proposed in this paper could also be used to model important influential nodes in social network analysis, which is an area that deserves further investigation.

\begin{acks}
This class file was developed by Sunrise Setting Ltd,
Brixham, Devon, UK.\\
Website: \url{http://www.sunrise-setting.co.uk}
\end{acks}

\bibliographystyle{apa}
\bibliography{references}

\section*{Appendix A: Description of Evaluation Measures}
\begin{itemize}
    \item True Positive (TP): True positives are instances classified as positive by the model that actually is positive. 
    \item True Negative (TN): True negatives are instances the model classifies as negative that actually are negative. 
    \item False Positive (FP): False positives are instances identified by the model as positive that actually are negative.
    \item False Negative (FN): False negatives are instances the model classifies as negative that actually are positive.  
    \item Accuracy (Ac): It is the percentage of correctly classified instances, and it is calculated as $\frac{TP + TN}{TP + TN + FP + FN}$.
    \item Precision (Pr): It calculates a model's ability to return only relevant instances. It is calculated as $\frac{TP}{TP + FP}$. 
    \item Recall (Re): It calculates a model's ability to identify all relevant instances. It is calculated as 
$\frac{TP}{TP + FN}$. 
    \item $F_1$ Score ($F_1$): A single metric that combines recall and precision using the harmonic mean. $F_1$ Score is calculated as $2 \times \frac{precision}{precision + recall}$.
    \item Cohen Kappa Score (CK Score):  Cohen's kappa score is used to measure inter-rater and itra-rater reliability for categorical items \citep{mchugh2012interrater}. It is calculated as $\frac{OA-AC}{1-AC}$, where $OA$ is the relative observed agreement between predicted labels and actual labels and $AC$ is the probability of agreement by chance. 
    \item Area Under Curve (AUC): The area under the Receiver operating characteristic (ROC) curve is called Area Under the Curve (AUC). ROC plots the true positive rate  versus the false positive rate as a function of the model’s threshold for classifying a positive. AUC calculates the overall performance of a classification model.
    %\item Micro Average (Micro avg):
    %\item Weighted Average (Weighted avg):
\end{itemize}

\end{document}